\documentclass{article}

    \PassOptionsToPackage{numbers, compress}{natbib}

    \usepackage[final]{neurips_2020_ml4ad}
\usepackage[utf8]{inputenc} 
\usepackage[T1]{fontenc}    
\usepackage{hyperref}       
\usepackage{url}            
\usepackage{booktabs}       
\usepackage{amsfonts}       
\usepackage{nicefrac}       
\usepackage{microtype}      

\usepackage{graphicx}
\usepackage{amsmath}
\usepackage{amssymb}
\usepackage{tabularx}
\usepackage{xmpmulti}
\usepackage{multirow}
\usepackage{arydshln}
\newcolumntype{L}[1]{>{\raggedright\let\newline\\\arraybackslash\hspace{0pt}}m{#1}}
\newcolumntype{C}[1]{>{\centering\arraybackslash}m{#1}}
\newcolumntype{R}[1]{>{\raggedleft\let\newline\\\arraybackslash\hspace{0pt}}m{#1}}
\usepackage{soul}

\title{SAFENet: Self-Supervised Monocular Depth Estimation with Semantic-Aware Feature Extraction}
\author{%
  Jaehoon Choi$^{\thanks{These two authors contributed equally.}}$\:\:\textsuperscript{\rm 1}, Dongki Jung$^{\footnotemark[1]}$\:\:\textsuperscript{\rm 1,2}, Donghwan Lee\textsuperscript{\rm 1}, Changick Kim\textsuperscript{\rm 2}\\
    \textsuperscript{\rm 1}NAVER LABS\\
    \textsuperscript{\rm 2}Korea Advanced Institute of Science and Technology\\
    \{jaehoon.c, donghwan.lee\}@naverlabs.com, \{jdk9405, changick\}@kaist.ac.kr}
\begin{document}

\maketitle

\begin{abstract}
Self-supervised monocular depth estimation has emerged as a promising method because it does not require groundtruth depth maps during training.
As an alternative for the groundtruth depth map, the photometric loss enables to provide self-supervision on depth prediction by matching the input image frames.
However, the photometric loss causes various problems, resulting in less accurate depth values compared with supervised approaches.
In this paper, we propose SAFENet that is designed to leverage semantic information to overcome the limitations of the photometric loss. 
Our key idea is to exploit semantic-aware depth features that integrate the semantic and geometric knowledge.
Therefore, we introduce multi-task learning schemes to incorporate semantic-awareness into the representation of depth features.
Experiments on KITTI dataset demonstrate that our methods compete or even outperform the state-of-the-art methods. Furthermore, extensive experiments on different datasets show its better generalization ability and robustness to various conditions, such as low-light or adverse weather.
\end{abstract}

\section{Introduction}
Monocular depth estimation which aims to perform dense depth estimation from a single image, is an important task in the field of autonomous driving, augmented reality, and robotics. Most supervised methods 
show that Convolutional Neural Networks (CNNs) are powerful tools to produce dense depth maps.
Nevertheless, collecting large-scale dense depth maps as groundtruth is significantly difficult because of data sparsity and expensive depth-sensing devices \cite{geiger2012we}, such as LiDAR. 
Accordingly, self-supervised monocular depth estimation \cite{garg2016unsupervised,zhou2017unsupervised} has gained significant attention in the recent years because it does not require image-groundtruth pairs. Self-supervised depth learning is a training method to regress the depth values via an error function, named the photometric loss, which computes the errors between the reference image and geometrically reprojected image from other viewpoints.
The reference image and the image from other viewpoints can be either a calibrated pair of left and right stereoscopic images\cite{garg2016unsupervised,godard2017unsupervised} 
or adjacent frames with the relative camera pose in a video sequence \cite{zhou2017unsupervised,godard2019digging}.    

\begin{figure}[t]
    \centering
    \includegraphics[width=0.8\linewidth]{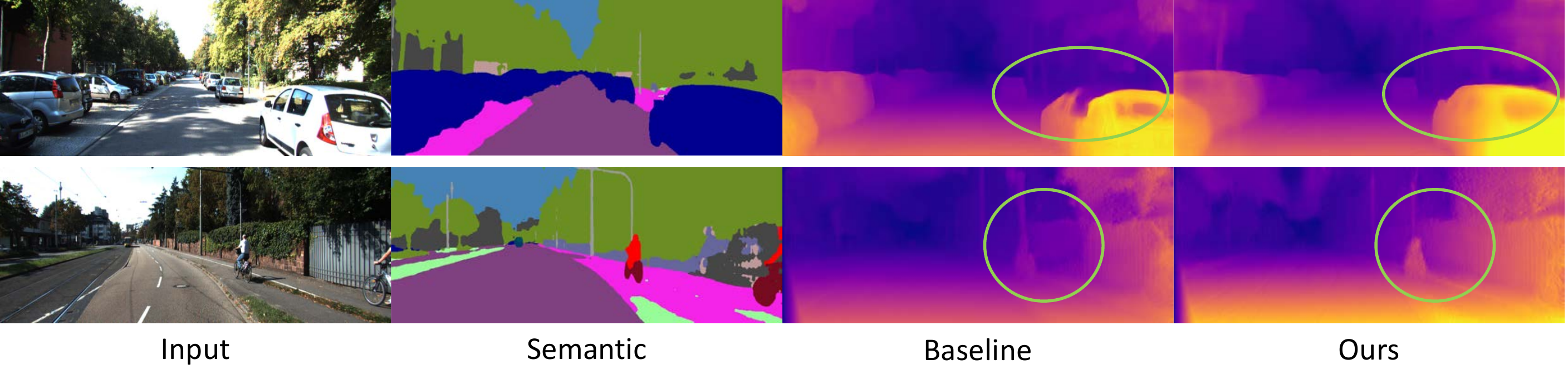}
    \vspace{-2mm}
    \caption{Example of monocular depth estimation based on self-supervision from monocular video sequences. The second column illustrates the improved depth prediction results by semantic-awareness.
    }
    \label{intro_figure}
    \vspace{-3mm}
\end{figure}

However, previous studies \cite{garg2016unsupervised,klodt2018supervising,godard2019digging} showed that the brightness change of pixels, low-textured regions, repeated patterns, and occlusions can cause differences in the photometric loss distribution and thus hinder the training. 
To address such limitations of the photometric loss, we propose a novel method that fuses the feature-level semantic information with geometric representations. Semantically-guided depth features might involve the spatial context of an input image.
This information (\textit{i.e.}, semantically-guided depth features)  serves as complementary knowledge in interpreting the three-dimensional (3D) Euclidean space, thereby improving the depth estimation performance. For example, from Fig. \ref{intro_figure}, it is evident that our method has a consistent depth range for each instance. In the first row, the distorted car shape of the baseline prediction is recovered in our prediction. However, despite these advantages, a general method of learning semantic-aware depth features has not been widely explored in the current self-supervised monocular depth estimation approaches. 

To learn semantic-aware depth features, we investigate multi-task learning (MTL) approaches that impose semantic supervision from the supervised segmentation task to self-supervised depth estimation task.
However, MTL often suffers from task interference, as the features learned to perform one task may not be appropriate to perform other tasks \cite{kokkinos2017ubernet}.
Thus, it is essential to distinguish the features between the task-specific and task-shared properties; \textit{i.e.}, one must know whether to share information for different tasks.

We propose a network architecture wherein two respective tasks share an encoder and have each decoder branch.
Task-specific schemes are designed to prevent corruption in the single encoder, and each subnetwork for the decoders contains task-sharing modules to establish a synergy between the tasks. In addition to these simple modules, we introduce a novel monocular depth estimation network that can consider the intermediate representation of semantic-awareness both in spatial and channel dimensions.

Our proposed strategy can be easily extended to both the types of self-supervised approaches: video sequences based and stereo images based. In this study, we focus on self-supervised learning from monocular video sequences. Furthermore, we experimentally validate the excellence of semantic-aware depth features under low-light and adverse weather conditions. The following are the contributions of this paper.
\begin{itemize}
    \item Novel approaches have been proposed to incorporate depth features with semantic features to perform self-supervised monocular depth estimation.
    \item It is demonstrated that the obtained semantic-aware depth features can overcome the drawbacks of the photometric loss, thereby enhancing the monocular depth estimation performance of networks.
    \item Our method achieves state-of-the-art results on the KITTI dataset, and extensive experiments on Virtual KITTI and nuScenes demonstrate that our method is more robust to various adverse conditions and better generalization capability than the current methods.
\end{itemize}

\section{Related Work}
\subsection{Self-supervised Monocular Depth Estimation}
Supervised monocular depth estimation models
\cite{eigen2014depth,laina2016deeper,chen2016single} 
require a large-scale groundtruth dataset, which is not only expensive to collect and but also has different characteristics depending on the sensors. To mitigate this issue, \cite{garg2016unsupervised} and \cite{godard2017unsupervised} proposed self-supervised training methods with stereo images. These methods exploited a warping function to transfer the coordinates of the left image to the right image plane. Simultaneously, instead of left-right consistency, \cite{zhou2017unsupervised} proposed a method to perform monocular depth estimation through camera ego-motion derived from video sequence images. This method computed the photometric loss by reprojecting adjacent frames to the current frame with the predicted depth and relative camera pose. Monodepth2 \cite{godard2019digging} enhanced the depth estimation performance using techniques such as the minimum reprojection error and auto-masking. Multiple studies relied on the assumption that image frames comprise rigid scenes, \textit{i.e.}, the appearance change in the spatial context is caused by the camera motion. Therefore, \cite{zhou2017unsupervised} applied network-predicted masks to moving objects, and \cite{godard2019digging} computed the per-pixel loss to ignore the regions where this assumption was violated. Additionally, to improve the quality of regression, many studies were conducted using additional cues, such as optical flow \cite{luo2018every,yin2018geonet,ranjan2019competitive} and edges  \cite{yang2018lego}. Recently, the methods in \cite{bian2019unsupervised,chen2019self} utilized geometric constraints as well as the photometric loss.

\subsection{Semantic Supervision}
Although semantic supervision is helpful for self-supervised monocular depth estimation, to the best of our knowledge, it has been discussed in only a few works. For performing self-supervision using stereo image pairs, \cite{ramirez2018geometry} utilized a shared encoder but separate decoders to jointly train both the tasks. 
\cite{chen2019towards} designed a left-right semantic consistency and semantics-guided smoothness regularization showing that semantic understanding increased the depth prediction accuracy. For video sequence models, some previous works \cite{casser2019depth,meng2019signet} also utilized information from either semantic- or instance-segmentation masks for the moving objects in the frames. The concurrent works \cite{zhu2020edge,klingner2020self} also presented a method to explicitly consider the relationship between depth estimation and semantic segmentation through either morphing operation or semantic masking for dynamic objects. The method in the recent work \cite{Guizilini2020Semantically-Guided}  is moderately similar to our method in that they both generated semantically-guided depth features by utilizing a fixed pretrained semantic network. However, instead of fixed semantic features, we adopt an end-to-end multi-task learning approach for performing monocular depth estimation.

\section{Proposed Approach}
\subsection{Motivation}
In this section, we discuss the mechanism of the photometric loss and limitations thereof. Additionally, we explain the reason of choosing semantic supervision to overcome the problems associated with the photometric loss.
\medskip
\newline
\textbf{Photometric Loss for Self-supervision.}\: 
Self-supervised monocular depth estimation relies on the photometric loss through warping between associated images, $I_{m}$ and $I_{n}$. These two images are sampled from the left-right pair in stereo vision or the adjacent time frames in a monocular video sequence. The photometric loss is formulated as follows:
\begin{equation}
\label{eq1} 
\begin{split}
L_{photo} & = \frac{1}{N}\sum_{p\in N}(\alpha \frac{1-\text{SSIM}_{mn}(p)}{2} + (1-\alpha)\parallel I_{m}(p)-I^\prime_{m}(p) \parallel)\:,
\end{split}
\end{equation}
where $I^\prime_{m}$ denotes the image obtained by warping $I_{n}$ with the predicted depth, \textit{N} the number of valid points successfully projected, and $\alpha$ is 0.85. In the case of video sequence model, the estimated camera pose and the intrinsic parameters are included in the warping process. However, the photometric loss has a severe drawback in that depth regression from RGB images is vulnerable to environmental changes. We hypothesize that the depth features jointly trained by semantic segmentation, called semantic-aware depth features, can leverage the semantic knowledge to assist in depth estimation. Therefore, we propose semantic supervision to resolve the issues of the photometric loss through multi-task learning. In the paper, our method mainly handles monocular video sequences but it can be globally adjusted to self-supervised networks regardless of stereo or sequence input.
For more details, please refer to the appendix.
\medskip
\newline
\textbf{Semantic Supervision.}\: 
Semantic-awareness can provide prior knowledge that if certain 3D points are projected to adjacent pixels with the same semantic class, then those points should be located at similar positions in the 3D space. Additionally, even in the regions where the RGB values are indistinguishable, understanding the spatial context using the semantic information can help comprehend the individual characteristics of the pixels in that region. To guide geometric reconstruction by the feature-level fusion of semantics, we design a method of learning two tasks through joint training rather than simply using segmentation masks as input.
For the supervised framework in the semantic segmentation task, pretrained DeepLabv3+ \cite{chen2018encoder} is used to prepare the pseudo labels of the semantic masks.

\subsection{Network Architecture}
Without a direct association between tasks, task interference might occur, which can corrupt each task-specific feature. Therefore, we present modules to obtain semantic-aware depth features by taking only those portions of the semantic features that are helpful for performing accurate depth estimation.
\begin{figure}[t]
\begin{center}
    \includegraphics[width=0.5\linewidth]{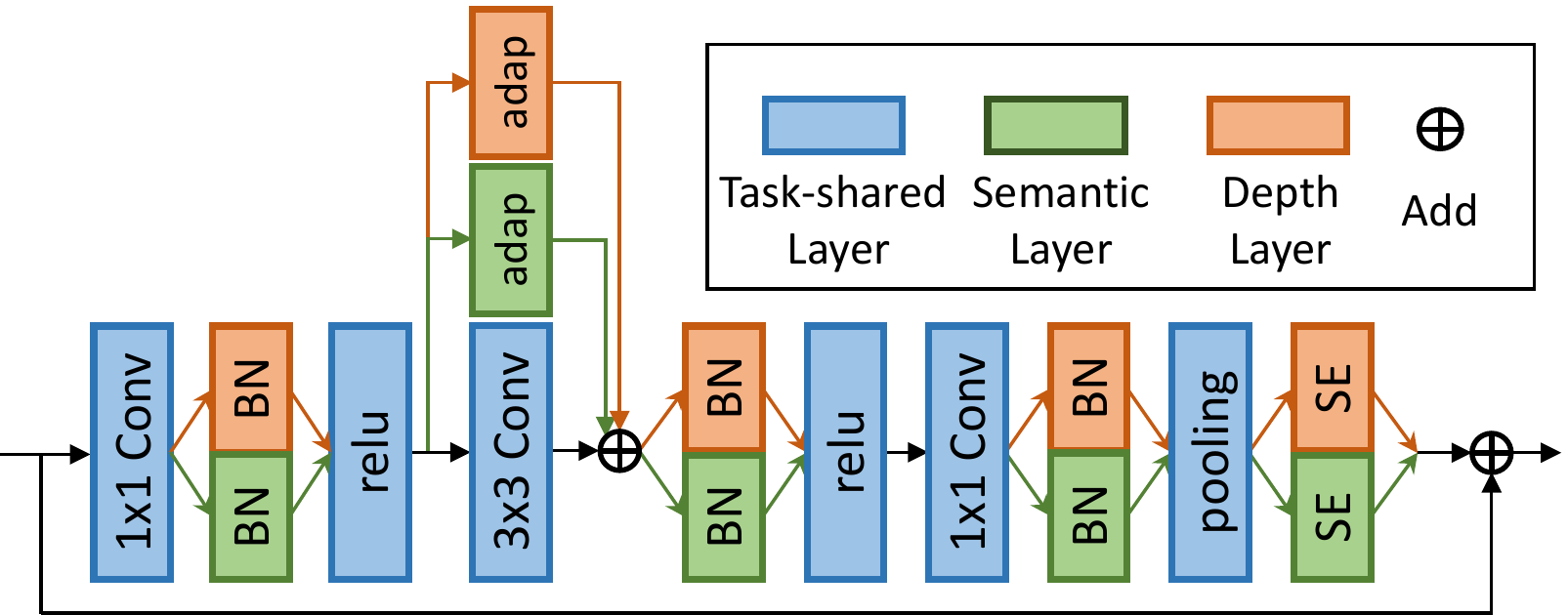}
\end{center}
\caption{SE-ResNet module for our encoder. 
    The terms ``SE'' and ``adapt'' denote the SE block \cite{hu2018squeeze} per task and task-specific residual adapter (RA) \cite{rebuffi2018efficient}, respectively.
    }
\label{figure2}
\end{figure}
\medskip
\newline
\textbf{Encoder.}\: To avoid interference between the depth estimation and segmentation, we build an encoder using three techniques of \cite{maninis2019attentive}, as shown in Fig. \ref{figure2}. First, the squeeze and excitation (SE) block \cite{hu2018squeeze} inserts global average pooled features into a fully connected layer and generates activated vectors for each channel via a sigmoid function.
The vectors that pass through the SE modules are multiplied with the features and give attention to each channel.
We then allocate different task-dependent parameters to the SE modules so that these modules can possess distinct characteristics. Second, Residual Adapters (RA) \cite{rebuffi2018efficient}, which introduce a few extra parameters that can possess task-specific attributes and rectify the shared features, are added to the existing residual layers as follows:
\begin{equation} \label{eq2}
    \text{L}_{\text{T}}(x) = x + \text{L}(x) + \text{RA}_{\text{T}}(x)\:,
\end{equation}
where $x$ denotes the features and $\text{T}\in \{\text{Depth}, \text{Seg\}}$. Additionally, $\text{L}(\cdot)$ and $\text{RA}_{\text{T}}(\cdot)$ denote a residual layer and a task-specific RA of task T, respectively.
Third, we obtain task-invariant features through batch normalization per task by exploiting the calculated statistics, which have task-dependent properties \cite{chang2019domain}.
\begin{figure*}[t!]
\begin{center}
    \includegraphics[width=0.85\linewidth]{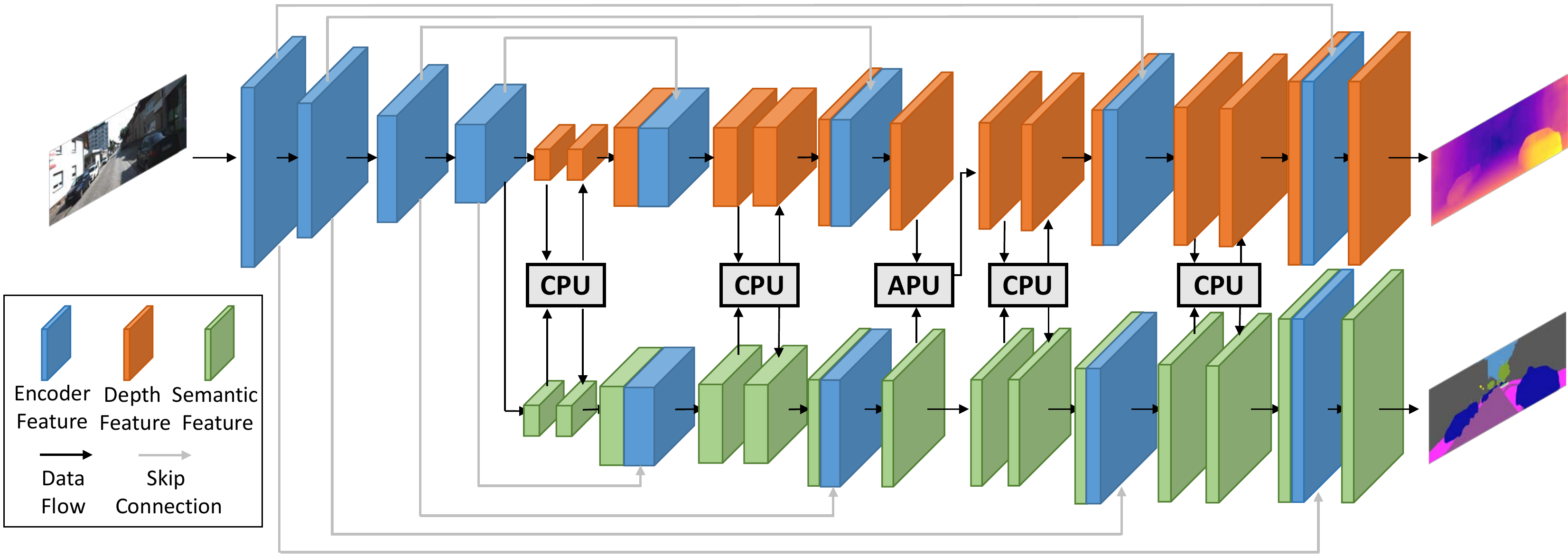}
    \includegraphics[width=0.85\linewidth]{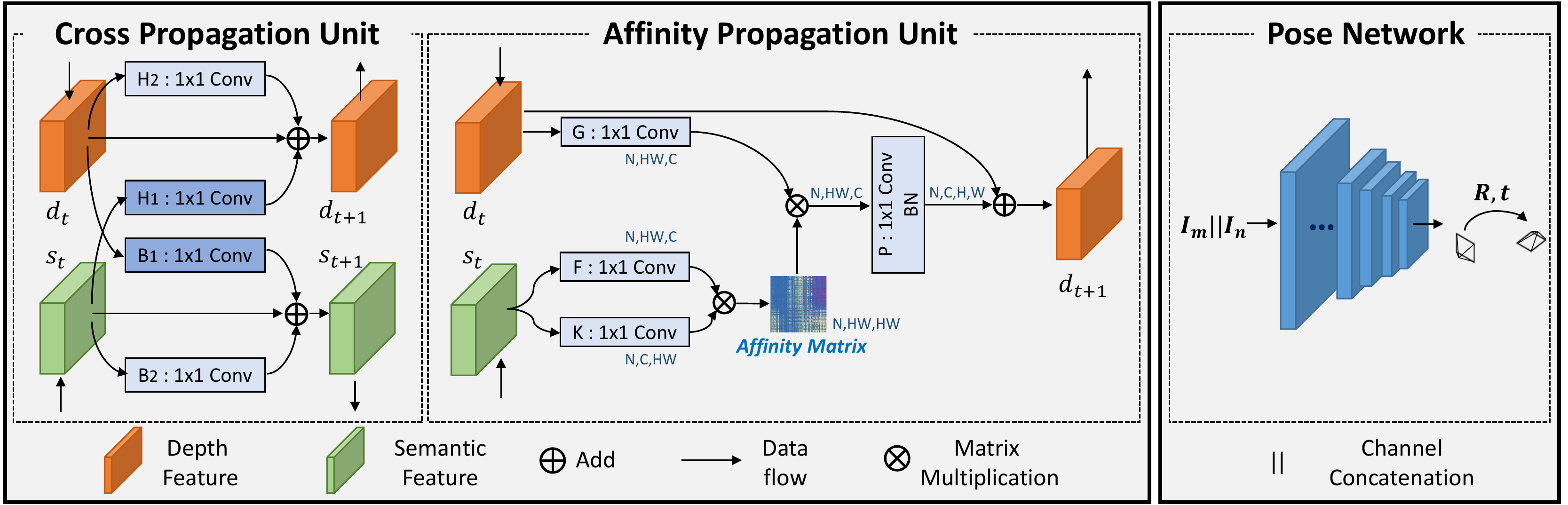}
\end{center}
  \vspace{-2mm}
   \caption{Overview of the proposed framework. In the top part, our network comprises one shared encoder and two separate decoders for each task. The bottom left part shows the proposed modules to propagate the information between two different tasks to learn semantic-aware depth features. The bottom right part denotes the pose estimation network. The details are provided in the appendix.} 
\label{Overview}
\vspace{-2mm}
\end{figure*}
\medskip
\newline
\textbf{Decoder.}\:
As shown in Fig. \ref{Overview}, we design a separate decoder for each task. Both the decoders can learn the task-specific features of their own, but find it difficult to exploit the features of the other decoder's task. We have experimented with two information propagation approaches to handle this issue. The first approach is inspired by the success of sharing units between two task networks in \cite{misra2016cross,jiao2018look}. Instead of weighted parameters suggested by previous works, we utilize 1\(\times\)1 convolutions \(\text{H}_{1}^{1\times1}(\cdot),\:\text{B}_{1}^{1\times1}(\cdot)\) to share the intermediate representations from the other task. Notably, both the 1\(\times\)1 convolutions, with the stride of 1, perform feature modulation only across channel dimensions. Before upsampling layers, we add \(\text{H}_{1}^{1\times1}(\cdot) ,\:\text{B}_{1}^{1\times1}(\cdot)\) enabling both the decoders to automatically share intermediate features rather than manually tuning the parameters for each feature. Also, we adopt a 1\(\times\)1 convolutional shortcut \(\text{H}_{2}^{1\times1}(\cdot),\:\text{B}_{2}^{1\times1}(\cdot)\) to reduce the negative effect of the interruption in propagation \cite{jiao2018look}, meaning that the features propagated from one task interfere with performing each other task. Given a segmentation feature \(s_{t}\) and depth feature \(d_{t}\), the task-shared features \(s_{t+1}\) and \(d_{t+1}\) can be obtained as follows:     
\begin{equation} \label{eq3}
\begin{split}
d_{t+1} &= d_{t} + \text{H}^{1\times1}_{1}(s_{t}) + \text{H}^{1\times1}_{2}(d_{t}),\:\:\: 
s_{t+1} = s_{t} +  \text{B}^{1\times1}_{1}(d_{t}) + \text{B}^{1\times1}_{2}(s_{t})\:.
\end{split}
\end{equation}
We refer to this module as the cross propagation unit (CPU). The second approach is to propagate the semantic affinity information from segmentation to depth estimation. Because all the above-mentioned sharing units comprise 1\(\times\)1 convolutions, the depth decoder cannot fuse the features at different spatial locations or learn the semantic affinity information captured by the segmentation decoder. Thanks to the feature extraction capability of CNNs, the high-dimension features from the segmentation decoder are used to learn the semantic affinity information. To learn a non-local affinity matrix, we first feed segmentation feature \(s_{t}\) into two 1\(\times\)1 convolution layers \(\text{K}^{1\times1}\:(\cdot)\) and \(\text{F}^{1\times1}\:(\cdot)\), where \({\text{K}^{1\times1}\:(s_{t}),\: \text{F}^{1\times1}\:(s_{t})} \in \rm I\!R^{C \times H \times W}\). Here, H, W, and C denote the height, width, and number of channels of the feature. After reshaping them to \(\rm I\!R^{C \times HW}\), we perform a matrix multiplication between the transpose of \(\text{F}^{1\times1}\:(s_{t})\) and \(\text{K}^{1\times1}\:(s_{t})\). By applying the softmax function, the affinity matrix \(\textbf{A} \in \rm I\!R^{HW \times HW}\) can be formulated as follows:
\begin{equation} \label{eq4}
a_{j,i} = \frac{\text{exp}(\text{F}^{1\times1}(s_{t})^{\text{T}}_{\text{i}} \cdot \text{K}^{1\times1}(s_{t})_{\text{j}})}{\sum_{\text{i}=1}^{\text{HW}} \text{exp}(\text{F}^{1\times1}(s_{t})^{\text{T}}_{\text{i}} \cdot \text{K}^{1\times1}(s_{t})_{\text{j}})}\:,
\end{equation}
where \(a_{j,i}\) denotes the affinity-propagation value at location \(j\) from the \(i\)-th region, and $\text{T}$ the transpose operation. Different than a non-local block \cite{wang2018non}, the semantic affinity matrix obtained is propagated to the depth features to transfer a semantic correlation of pixel-wise features. We then conduct a matrix multiplication between the depth features from \(\text{G}^{1\times1}(\cdot)\) and semantic affinity matrix \textbf{A}. Subsequently, we can obtain depth features guided by the semantic affinity matrix. To mitigate the interruption in propagation \cite{jiao2018look}, we add the original depth feature to the result of affinity propagation. The affinity-propagation process can be expressed as follows: 
\begin{equation} \label{eq5}
d_{t+1} = BN(\text{P}^{1\times1}(\textbf{A}\text{G}^{1\times1}(d_t))) + d_t\:,
\end{equation}
where \(\text{P}^{1\times1}\) and \(BN\) denote a 1\(\times\)1 convolution layer and batch normalization layer, respectively. This module is named the affinity propagation unit (APU). This spatial correlation of semantic features is critical to accurately estimate the depth in the self-supervised regime. 

\subsection{Loss Functions}
Our loss function comprises supervised and self-supervised loss terms. For semantic supervision, pseudo labels or groundtruth annotations are available. We define the semantic segmentation loss \(L_{seg}\) using cross entropy. As previously described, we use the phtometric loss \(L_{photo}\) in Eq. (\ref{eq1}) for self-supervised training. 
Additionally, to regularize the depth in a low texture or homogeneous region of the scene,  we adopt the edge-aware depth smoothness loss \(L_{smooth}\) in \cite{godard2017unsupervised}.   

Consequently, the overall loss function is formulated as follows, 
\begin{equation}\label{eq6}
    L_{tot} = L_{photo} + \lambda_{smooth} L_{smooth} + \lambda_{seg} L_{seg}\:,
\end{equation}
where \(\lambda_{seg}\) and \(\lambda_{smooth}\) denote weighting terms selected through grid search. Notably, our network can be trained in an end-to-end manner. All the parameters in the encoder's task-shared modules, APU and CPU are trained by the back-propagation of \(L_{tot}\), while the parameters in the task-specific modules of the encoder and decoders are learned by the gradient of the task-specific loss, namely \(L_{seg}\) or \(L_{photo} + L_{smooth}\). For instance, all the specific layers for the segmentation task both in the encoder and decoder are not trained with \(L_{photo}\) and \(L_{smooth}\), and vice versa.

Furthermore, for performing self-supervised training using a monocular video sequence, we simultaneously train an additional pose network and the proposed encoder-decoder model. The pose network follows the training protocols described in Monodepth2 \cite{godard2019digging}. We also incorporate the techniques in Monodepth2, including auto-masking, applying per-pixel minimum reprojection loss, and depth map upsampling to obtain improved results. 

\section{Experiments}
In this section, we evaluate the proposed approach on performing self-supervised monocular depth estimation using monocular video sequences. We also compare the proposed approach with other state-of-the-art methods.

\begin{table*}[t]
    \centering
    \resizebox{0.95\textwidth}{!}{\normalsize\begin{tabular}{l||cccc|ccc}
    \hline
    \multirow{2}{*}{Method} & \multicolumn{4}{c|}{Lower is better.} & \multicolumn{3}{c}{Higher is better.} \\ 
    & Abs Rel & Sq Rel\: & RMSE & RMSE log & $\delta<1.25$ & $\delta<1.25^2$ & $\delta<1.25^3$\\ \hline
    Zhou \cite{zhou2017unsupervised}* & 0.183 & 1.595 & 6.709 & 0.270 & 0.734 & 0.902 & 0.959 \\
    LEGO \cite{yang2018lego} & 0.162 & 1.352 & 6.276 & 0.252 & - & - & - \\
    GeoNet \cite{yin2018geonet}* & 0.149 & 1.060 & 5.567 & 0.226 & 0.796 & 0.935 & 0.975 \\
    DF-Net \cite{zou2018df} & 0.150 & 1.124 & 5.507 & 0.223 & 0.806 & 0.933 & 0.973 \\
    EPC++ \cite{luo2018every} & 0.141 & 1.029 & 5.350 & 0.216 & 0.816 & 0.941 & 0.976 \\
    Struct2depth \cite{casser2019depth} & 0.141 & 1.026 & 5.291 & 0.215 & 0.816 & 0.945 & 0.979 \\
    SC-SfMLearner\cite{bian2019unsupervised} & 0.137 & 1.089 & 5.439 & 0.217 & 0.830 & 0.942 & 0.975 \\
    CC \cite{ranjan2019competitive} & 0.140 & 1.070 & 5.326 & 0.217 & 0.826 & 0.941 & 0.975 \\
    SIGNet \cite{meng2019signet} & 0.133 & 0.905 & 5.181 & 0.208 & 0.825 & 0.947 & 0.981 \\
    GLNet \cite{chen2019self} & 0.135 & 1.070 & 5.230 & 0.210 & 0.841 & 0.948 & 0.980 \\
    Monodepth2 \cite{godard2019digging} & 0.115 & 0.903 & 4.863 & 0.193 & 0.877 & 0.959 & 0.981 \\
    Guizilini, ResNet18 \cite{Guizilini2020Semantically-Guided} & 0.117 & 0.854 & 4.714 & 0.191 & 0.873 & \bf{0.963} & 0.981 \\
    Johnston, ResNet101 \cite{johnston2020self} & \bf{0.106} & 0.861 & 4.699 & \bf{0.185} & \bf{0.889} & 0.962 & 0.982 \\
    SGDepth, ResNet18 \cite{klingner2020self} & 0.113 & 0.835 & 4.693 & 0.191 & 0.879 & 0.961 & 0.981 \\
    \textbf{SAFENet} $(640\times192)$ & 0.112 & \bf{0.788} & \bf{4.582} & 0.187 & 0.878 & \bf{0.963} & \bf{0.983}\\ \hline
    \textbf{SAFENet} $(1024\times320)$ & \bf{0.106} & \bf{0.743} & \bf{4.489} & \bf{0.181} & 0.884 & \bf{0.965} & \bf{0.984}\\
    \hline
    \end{tabular}}
    \caption{Quantitative results on the KITTI 2015 dataset \cite{geiger2012we} by using the split of Eigen. * indicates updated results from GitHub. 
    We additionally achieved better performance under the high resolution $1024\times320$.
    This table does not include online refinement performance for ensuring a fair comparison.}
    \label{table1}
\end{table*}
\begin{table*}[!t]
    \centering
    \resizebox{0.85\textwidth}{!}{\normalsize\begin{tabular}{l||cccc|ccc}
    \hline
    \multirow{2}{*}{Method} & \multicolumn{4}{c|}{Lower is better.} & \multicolumn{3}{c}{Higher is better.} \\ 
    & Abs Rel & Sq Rel\: & RMSE & RMSE log & $\delta<1.25$ & $\delta<1.25^2$ & $\delta<1.25^3$ \\ \hline
    Monodepth2 \cite{godard2019digging} & 0.187 & 1.865 & 8.322 & 0.303 & 0.722 & 0.882 & 0.939 \\
    \textbf{SAFENet} $(640\times192)$ & \bf{0.172} & \bf{1.652} & \bf{7.776} & \bf{0.277} & \bf{0.752} & \bf{0.895} & \bf{0.950} \\ \hline
    \textbf{SAFENet} $(1024\times320)$ & 0.175 & 1.667 & \bf{7.533} & \bf{0.274} & 0.750 & \bf{0.902} & \bf{0.951} \\
    \hline
    \end{tabular}}
    \caption{Quantitative results on the nuScenes dataset \cite{caesar2020nuscenes}
    }
    \label{nuscenes}
    \vspace{-5mm}
\end{table*}

\subsection{Experimental Settings}
\textbf{KITTI.}\:
We used the KITTI dataset \cite{geiger2012we} as in \cite{zhou2017unsupervised}. The dataset comprises 39,810 triple frames for training and 4,424 images for validation in the video sequence model. The test split comprises 697 images. Because these images had no segmentation labels, we prepared semantic masks of 19 categories from DeepLabv3+ pretrained on Cityscapes \cite{cordts2016cityscapes}. The pretrained model attained the segmentation performance of mIoU 75\% on the KITTI validation set.
\newline
\textbf{Virtual KITTI.}\:
To demonstrate that our method performs robustly in the adverse weather, we experimented with the Virtual KITTI (vKITTI) dataset \cite{gaidon2016virtual}, a synthetic dataset comprising various weather conditions in five video sequences and 11 classes of semantic labels. We then divided the vKITTI dataset on the basis of six weather conditions, as described in \cite{gaidon2016virtual}. The training set had relatively clean 8,464 sequence triplets that belonged to morning, sunset, overcast, and clone images. 4,252 fog and clone images, which are challenging because of having environments significantly different than those of the images in the training set, were tested to show each performance.
\newline
\textbf{nuScenes.}\:
The nuScenes-mini comprises 404 front-camera images of 10 different scenes, and the corresponding depth labels from LiDAR sensors. To evaluate the generalization for other types of images from other datasets, we applied models pretrained with KITTI to nuScenes without fine-tuning.
All the predicted depth ranges on the KITTI, vKITTI, and nuScenes were clipped to 80 m to match the Eigen via following \cite{godard2019digging}.
\medskip
\newline
\textbf{Implementation Details.}\:
We built our encoder based on the ResNet-18 \cite{he2016deep} backbone with SE modules, and bridged the encoder to the decoder with skip connections based on the general U-Net architecture. 
Each layer of the encoder was pretrained on the ImageNet dataset 
while the parameters in the task-specific modules of the encoder, and two decoders, CPU and APU were randomly initialized. We used a ResNet based pose network following Monodepth2 \cite{godard2019digging}.       

\subsection{Experimental Results}
\textbf{Comparison with State-of-the-art Methods.}\:
The quantitative results of self-supervised monocular depth estimation on the KITTI dataset are presented in Table \ref{table1}.
Our method outperformed not only the baseline \cite{godard2019digging} but also other networks in terms of most of the metrics.
Conversely, the limitation of the photometric loss, which compares individual errors at the pixel level, can be improved by supervision from feature-level semantic information.
In Table \ref{nuscenes}, we have evaluated the generalization capability of our methods on the nuScenes dataset.
More number of qualitative results are provided in the appendix.
\begin{table*}[t]
    \centering
    \resizebox{0.9\textwidth}{!}{\normalsize\begin{tabular}{l|c||cccc|ccc}
    \hline
    \multirow{2}{*}{Method} & \multirow{2}{*}{Weather} & \multicolumn{4}{c|}{Lower is better.} & \multicolumn{3}{c}{Higher is better.} \\ 
    & & Abs Rel & Sq Rel\: & RMSE & RMSE log & $\delta<1.25$ & $\delta<1.25^2$ & $\delta<1.25^3$ \\ \hline
    Monodepth2 \cite{godard2019digging} (SE) & fog & 0.218 & 2.823 & 10.392 & 0.370 & 0.686 & 0.871 & 0.919\\
    \textbf{SAFENet} & fog & \bf{0.213} & \bf{2.478} & \bf{9.018} & \bf{0.317} & \bf{0.690} & \bf{0.872} & \bf{0.936} \\ \hline
    Monodepth2 \cite{godard2019digging} (SE) & rain & 0.200 & 1.907 & 6.965 & 0.263 & 0.734 & 0.901 & 0.961 \\
    \textbf{SAFENet} & rain & \bf{0.145} & \bf{1.114} & \bf{6.349} & \bf{0.222} & \bf{0.800} & \bf{0.937} & \bf{0.977} \\
    \hline
    \end{tabular}}
    \caption{Adverse weather experiments on the vKITTI dataset \cite{gaidon2016virtual}. For ensuring a fair comparison, we test after adding SE modules to the base architecture of Monodepth2.}
    \label{vkitti_weather}
\end{table*}
\begin{table*}[t]
    \centering
    \resizebox{0.85\textwidth}{!}{\normalsize\begin{tabular}{l|cccc||cccc|ccc}
    \hline
    \multirow{2}{*}{Model} & \multirow{2}{*}{Seg} & \multirow{2}{*}{R/N}&  \multirow{2}{*}{CPU} &  \multirow{2}{*}{APU} & \multicolumn{4}{c|}{Lower is better.} & \multicolumn{3}{c}{Higher is better.} \\ 
    & & & & & Abs Rel & Sq Rel\: & RMSE & RMSE log \: &\:  $\delta<1.25$ & $\delta<1.25^2$ & $\delta<1.25^3$ \\ 
    \hline
    Monodepth2 &  &  &  &  & 0.115 & 0.903 & 4.863 & 0.193 & 0.877 & 0.959 & 0.981 \\
    \hline
    SAFENet & \checkmark  & & & & 0.116 & 0.918 & 4.842 & 0.193 & 0.873 & 0.959 & 0.981 \\
    SAFENet & \checkmark  & \checkmark & & &  0.116 & 0.883 & 4.703 & 0.189 & 0.877 & 0.961 & 0.982 \\
    SAFENet & \checkmark  & \checkmark & \checkmark & &  0.117 & 0.826 & 4.660 & \bf{0.187} & 0.869 & 0.961 & \bf{0.984}\\
    SAFENet & \checkmark  & \checkmark &  & \checkmark &  \bf{0.111} & 0.815 & 4.665 & \bf{0.187} & \bf{0.881} & 0.962 & 0.982\\
    \textbf{SAFENet} & \checkmark  & \checkmark & \checkmark & \checkmark &  0.112 & \bf{0.788} & \bf{4.582} & \bf{0.187} & 0.878 & \bf{0.963} & 0.983\\
    \hline
    \end{tabular}}
    \caption{Ablation for the proposed model. Seg is multi-task learning with segmentation. The term R and N denote the task-specific RA and batch normalization per task, respectively.
    }
    \label{monodepth2_ablation}
    \vspace{-4mm}
\end{table*}

\begin{figure*}[t]
\begin{center}
    \centering
    \includegraphics[width=0.8\linewidth]{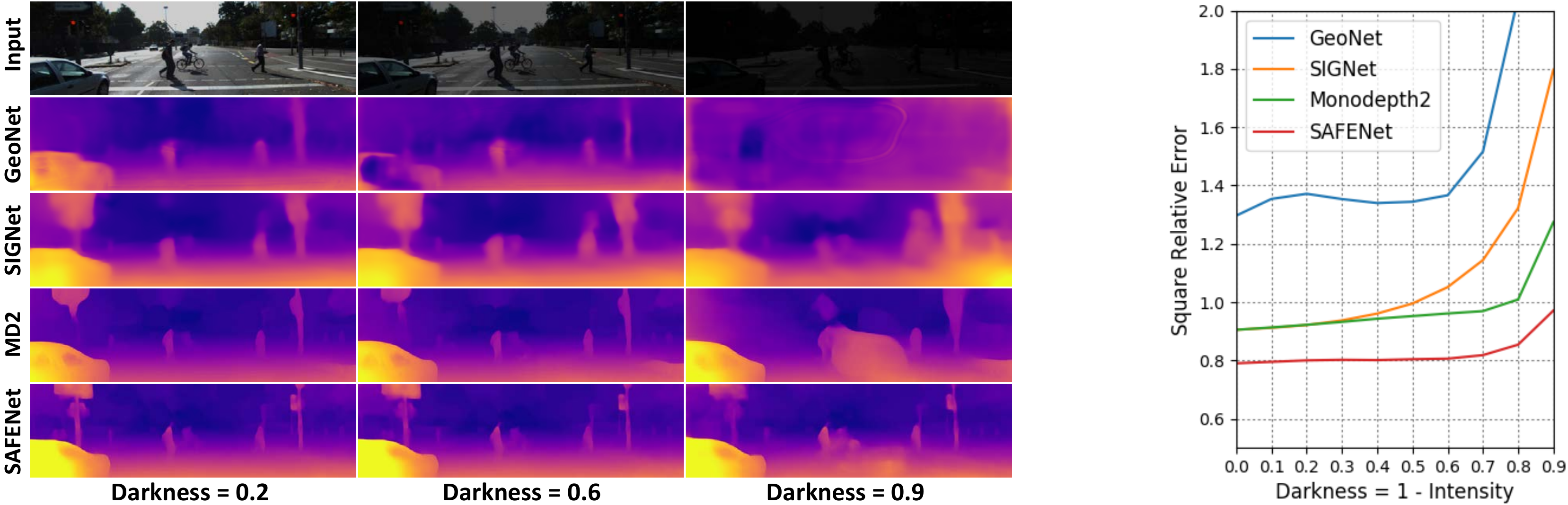}
    \caption{Robustness to changes in the light intensity. The qualitative results from the top to bottom show the input and depth predictions of GeoNet \cite{yin2018geonet}, SIGNet \cite{meng2019signet}, Monodepth2 \cite{godard2019digging}, and SAFENet. In the graph, we show the most steady square relative errors irrespective of the light intensity. }
    \vspace{-2mm}
    \label{fig5}
\end{center}
\end{figure*}

\begin{figure*}[!t]
    \centering
    \includegraphics[width=0.8\linewidth]{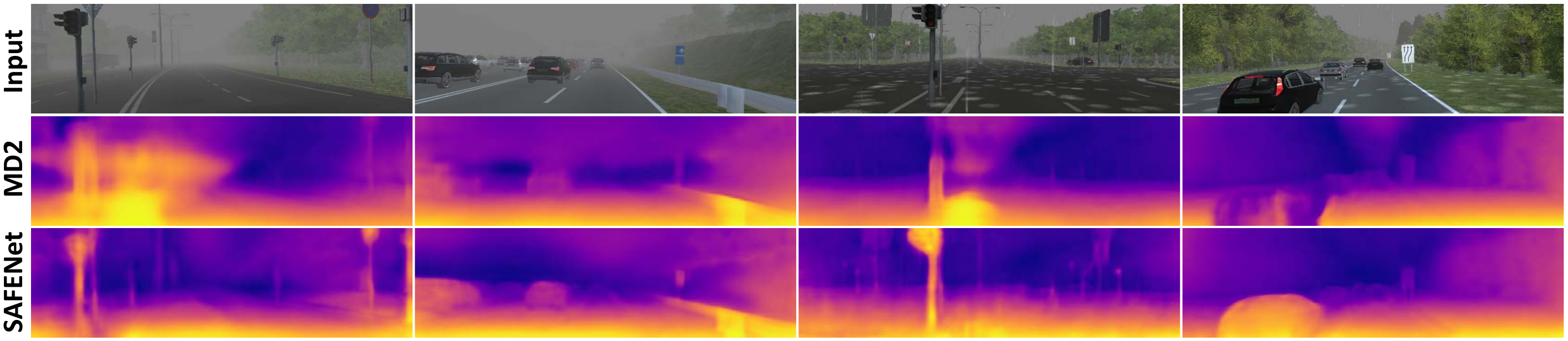}
    \caption{Qualitative results on the fog and rain data of the vKITTI dataset \cite{gaidon2016virtual}. 
    The left two images are of fog conditions, and the right two ones are of rainy conditions.}
    \label{weather_visual}
    \vspace{-2mm}
\end{figure*}

\medskip
\textbf{Low Light Conditions.}\:
Assuming low light situations, we measured the performance of the networks by multiplying the input images by a scale that ranged between zero and one. Figure \ref{fig5} shows that our proposed method achieved consistent results irrespective of the illuminance-level. When darkness value becomes 0.9, compared with other methods, our approach exhibited a smaller increase in the square relative error. This proves that our method complements depth estimation by identifying semantics rather than simply regressing the depth values from RGB information.
\medskip
\newline
\textbf{Weather Conditions.}\:
In addition to the low light experiments, we experiment on the vKITTI dataset to show that the proposed method is robust to the adverse weather. After training with the data of other conditions, we tested the cases of rain and fog, both of which are challenging for depth estimation. From Table \ref{vkitti_weather}, it is evident that the performance of the proposed method improved when depth estimation was performed using semantic-aware depth features. Correspondingly, Fig. \ref{weather_visual} shows that the problems associated with depth hole (first column) or infinite depth on moving objects (fourth column) are reduced, and the shape of the objects is thus satisfactorily predicted.

\begin{table*}[h]
    \centering
    \resizebox{0.35\textwidth}{!}{\begin{tabular}{l|c|c|c}
    \hline
     & \small{\cite{chen2019towards}} &  \small{\cite{klingner2020self}}  & \textbf{SAFENet} \\
    \hline \hline
    mIoU$_{K}$ & 37.7 & 51.6 & \textbf{61.2}  \\
    \hline
    \end{tabular}}
    \caption{Evaluation of semantic segmentation on the KITTI 2015 split.}
    \label{segmentationresults}
\end{table*}

\begin{figure*}[!t]
    \centering
    \includegraphics[width=0.8\linewidth]{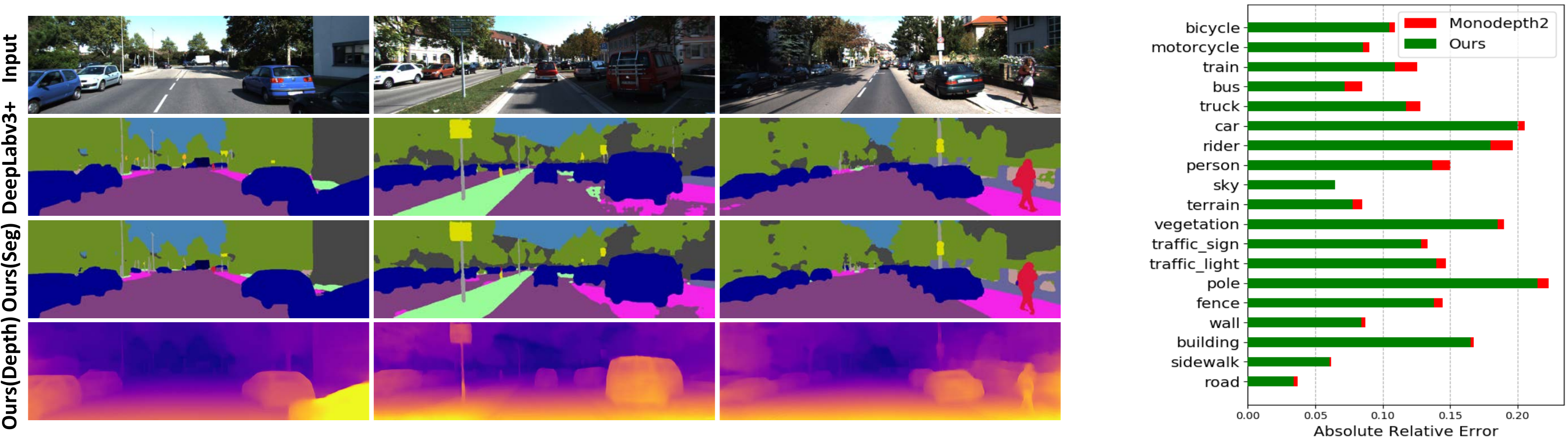}
    \vspace{-3mm}
    \caption{Comparison of depth estimation error among distinct classes. Our method improves the performance in all the classes except for the sky class, which has infinite depth.}
    \label{class-specific}
    \vspace{-2mm}
\end{figure*}

\textbf{Further Discussion about Semantic Supervision.}\:
Although this paper does not directly address the semantic segmentation task, the segmentation accuracy can provide a better understanding of our method. Our method perform notably compared with other methods in Table. \ref{segmentationresults}. Through the aforementioned experiments, we demonstrated that our training schemes are sufficient to present geometric features with semantic-awareness. However, we showed segmentation results only on KITTI split. Because our method exploits Cityscapes to pretrain pseudo label generator, training with Cityscapes conflicts with our experimental setting. To demonstrate the strength of semantic-aware depth features, the performance results on each class are shown in Fig. \ref{class-specific}. We have exploited semantic masks per class to evaluate the class-specific depth estimation performance. Using semantic information, our method shows that the absolute relative difference is reduced in all the classes except for the sky class. Particularly, people (0.150 to 0.137) and poles (0.223 to 0.215) have performance improvement. The accurate depth values of these categories are difficult to learn by the photometric loss because of their exquisite shapes. However, the semantic-aware features satisfactorily delineate the contours of the objects. Besides, it is seen that semantic-awareness is also helpful for estimating the distances of moving classes, such as the riders (0.197 to 0.180) and trains (0.125 to 0.109), which violate the assumption of rigid motion in self-supervised monocular depth training.


\medskip
\textbf{Ablation Study.}\:
We conducted experiments to explore the effects of the proposed methods while removing each module in Table. \ref{monodepth2_ablation}. Significant improvement occured in almost all the metrics when semantic-aware depth features were created by using our techniques, which divide task-specific and task-shared parameters. CPU and APU process the features in the channel and spatial dimensions, respectively, and achieve better results when both of them are included in the networks. In the appendix, we provided the ablation studies on depth estimation trained via stereo vision.
\section{Conclusions}
We discussed the problems of the photometric loss and introduced ways solve those problems using semantic information. Through the designed multi-task approach, our self-supervised depth estimation network could learn semantic-aware features to improve the depth prediction performance.
The proposed modules could be universally applied to self-supervision depth networks. Furthermore, to prove the robustness of our method to environmental changes, various experiments were conducted under different conditions. The experimental results showed that our method was more effective than other state-of-the-art methods.



\bibliographystyle{abbrv}
\bibliography{bib}

\begin{thebibliography}{10}

\bibitem{bian2019unsupervised}
J.~Bian, Z.~Li, N.~Wang, H.~Zhan, C.~Shen, M.-M. Cheng, and I.~Reid.
\newblock Unsupervised scale-consistent depth and ego-motion learning from
  monocular video.
\newblock In {\em NeurIPS}, pages 35--45, 2019.

\bibitem{caesar2020nuscenes}
H.~Caesar, V.~Bankiti, A.~H. Lang, S.~Vora, V.~E. Liong, Q.~Xu, A.~Krishnan,
  Y.~Pan, G.~Baldan, and O.~Beijbom.
\newblock nuscenes: A multimodal dataset for autonomous driving.
\newblock In {\em CVPR}, pages 11621--11631, 2020.

\bibitem{casser2019depth}
V.~Casser, S.~Pirk, R.~Mahjourian, and A.~Angelova.
\newblock Depth prediction without the sensors: Leveraging structure for
  unsupervised learning from monocular videos.
\newblock In {\em AAAI}, volume~33, pages 8001--8008, 2019.

\bibitem{chang2019domain}
W.-G. Chang, T.~You, S.~Seo, S.~Kwak, and B.~Han.
\newblock Domain-specific batch normalization for unsupervised domain
  adaptation.
\newblock In {\em CVPR}, pages 7354--7362, 2019.

\bibitem{chen2018encoder}
L.-C. Chen, Y.~Zhu, G.~Papandreou, F.~Schroff, and H.~Adam.
\newblock Encoder-decoder with atrous separable convolution for semantic image
  segmentation.
\newblock In {\em ECCV}, pages 801--818, 2018.

\bibitem{chen2019towards}
P.-Y. Chen, A.~H. Liu, Y.-C. Liu, and Y.-C.~F. Wang.
\newblock Towards scene understanding: Unsupervised monocular depth estimation
  with semantic-aware representation.
\newblock In {\em CVPR}, pages 2624--2632, 2019.

\bibitem{chen2016single}
W.~Chen, Z.~Fu, D.~Yang, and J.~Deng.
\newblock Single-image depth perception in the wild.
\newblock In {\em NeurIPS}, pages 730--738, 2016.

\bibitem{chen2019self}
Y.~Chen, C.~Schmid, and C.~Sminchisescu.
\newblock Self-supervised learning with geometric constraints in monocular
  video: Connecting flow, depth, and camera.
\newblock In {\em ICCV}, pages 7063--7072, 2019.

\bibitem{cordts2016cityscapes}
M.~Cordts, M.~Omran, S.~Ramos, T.~Rehfeld, M.~Enzweiler, R.~Benenson,
  U.~Franke, S.~Roth, and B.~Schiele.
\newblock The cityscapes dataset for semantic urban scene understanding.
\newblock In {\em CVPR}, pages 3213--3223, 2016.

\bibitem{eigen2014depth}
D.~Eigen, C.~Puhrsch, and R.~Fergus.
\newblock Depth map prediction from a single image using a multi-scale deep
  network.
\newblock In {\em NeurIPS}, pages 2366--2374, 2014.

\bibitem{gaidon2016virtual}
A.~Gaidon, Q.~Wang, Y.~Cabon, and E.~Vig.
\newblock Virtual worlds as proxy for multi-object tracking analysis.
\newblock In {\em CVPR}, pages 4340--4349, 2016.

\bibitem{garg2016unsupervised}
R.~Garg, V.~K. BG, G.~Carneiro, and I.~Reid.
\newblock Unsupervised cnn for single view depth estimation: Geometry to the
  rescue.
\newblock In {\em ECCV}, pages 740--756. Springer, 2016.

\bibitem{geiger2012we}
A.~Geiger, P.~Lenz, and R.~Urtasun.
\newblock Are we ready for autonomous driving? the kitti vision benchmark
  suite.
\newblock In {\em CVPR}, pages 3354--3361. IEEE, 2012.

\bibitem{godard2017unsupervised}
C.~Godard, O.~Mac~Aodha, and G.~J. Brostow.
\newblock Unsupervised monocular depth estimation with left-right consistency.
\newblock In {\em CVPR}, pages 270--279, 2017.

\bibitem{godard2019digging}
C.~Godard, O.~Mac~Aodha, M.~Firman, and G.~J. Brostow.
\newblock Digging into self-supervised monocular depth estimation.
\newblock In {\em ICCV}, pages 3828--3838, 2019.

\bibitem{Guizilini2020Semantically-Guided}
V.~Guizilini, R.~Hou, J.~Li, R.~Ambrus, and A.~Gaidon.
\newblock Semantically-guided representation learning for self-supervised
  monocular depth.
\newblock In {\em ICLR}, 2020.

\bibitem{he2016deep}
K.~He, X.~Zhang, S.~Ren, and J.~Sun.
\newblock Deep residual learning for image recognition.
\newblock In {\em CVPR}, pages 770--778, 2016.

\bibitem{hirschmuller2005accurate}
H.~Hirschmuller.
\newblock Accurate and efficient stereo processing by semi-global matching and
  mutual information.
\newblock In {\em 2005 IEEE Computer Society Conference on Computer Vision and
  Pattern Recognition (CVPR'05)}, volume~2, pages 807--814. IEEE, 2005.

\bibitem{hu2018squeeze}
J.~Hu, L.~Shen, and G.~Sun.
\newblock Squeeze-and-excitation networks.
\newblock In {\em CVPR}, pages 7132--7141, 2018.

\bibitem{jaderberg2015spatial}
M.~Jaderberg, K.~Simonyan, A.~Zisserman, et~al.
\newblock Spatial transformer networks.
\newblock In {\em NeurIPS}, pages 2017--2025, 2015.

\bibitem{jiao2018look}
J.~Jiao, Y.~Cao, Y.~Song, and R.~Lau.
\newblock Look deeper into depth: Monocular depth estimation with semantic
  booster and attention-driven loss.
\newblock In {\em ECCV}, pages 53--69, 2018.

\bibitem{johnston2020self}
A.~Johnston and G.~Carneiro.
\newblock Self-supervised monocular trained depth estimation using
  self-attention and discrete disparity volume.
\newblock In {\em CVPR}, pages 4756--4765, 2020.

\bibitem{kingma2014adam}
D.~P. Kingma and J.~Ba.
\newblock Adam: A method for stochastic optimization.
\newblock {\em arXiv}, 2014.

\bibitem{klingner2020self}
M.~Klingner, J.-A. Term{\"o}hlen, J.~Mikolajczyk, and T.~Fingscheidt.
\newblock Self-supervised monocular depth estimation: Solving the dynamic
  object problem by semantic guidance.
\newblock {\em arXiv}, 2020.

\bibitem{klodt2018supervising}
M.~Klodt and A.~Vedaldi.
\newblock Supervising the new with the old: learning sfm from sfm.
\newblock In {\em ECCV}, pages 698--713, 2018.

\bibitem{kokkinos2017ubernet}
I.~Kokkinos.
\newblock Ubernet: Training a universal convolutional neural network for low-,
  mid-, and high-level vision using diverse datasets and limited memory.
\newblock In {\em CVPR}, pages 6129--6138, 2017.

\bibitem{laina2016deeper}
I.~Laina, C.~Rupprecht, V.~Belagiannis, F.~Tombari, and N.~Navab.
\newblock Deeper depth prediction with fully convolutional residual networks.
\newblock In {\em 3DV}, pages 239--248. IEEE, 2016.

\bibitem{luo2018every}
C.~Luo, Z.~Yang, P.~Wang, Y.~Wang, W.~Xu, R.~Nevatia, and A.~Yuille.
\newblock Every pixel counts++:joint learning of geometry and motion with 3d
  holistic understanding.
\newblock {\em arxiv}, 2018.

\bibitem{maninis2019attentive}
K.-K. Maninis, I.~Radosavovic, and I.~Kokkinos.
\newblock Attentive single-tasking of multiple tasks.
\newblock In {\em CVPR}, pages 1851--1860, 2019.

\bibitem{meng2019signet}
Y.~Meng, Y.~Lu, A.~Raj, S.~Sunarjo, R.~Guo, T.~Javidi, G.~Bansal, and
  D.~Bharadia.
\newblock Signet: Semantic instance aided unsupervised 3d geometry perception.
\newblock In {\em CVPR}, pages 9810--9820, 2019.

\bibitem{misra2016cross}
I.~Misra, A.~Shrivastava, A.~Gupta, and M.~Hebert.
\newblock Cross-stitch networks for multi-task learning.
\newblock In {\em CVPR}, pages 3994--4003, 2016.

\bibitem{poggi2018learning}
M.~Poggi, F.~Tosi, and S.~Mattoccia.
\newblock Learning monocular depth estimation with unsupervised trinocular
  assumptions.
\newblock In {\em 3DV}, pages 324--333. IEEE, 2018.

\bibitem{ramirez2018geometry}
P.~Z. Ramirez, M.~Poggi, F.~Tosi, S.~Mattoccia, and L.~Di~Stefano.
\newblock Geometry meets semantics for semi-supervised monocular depth
  estimation.
\newblock In {\em ACCV}, pages 298--313. Springer, 2018.

\bibitem{ranjan2019competitive}
A.~Ranjan, V.~Jampani, L.~Balles, K.~Kim, D.~Sun, J.~Wulff, and M.~J. Black.
\newblock Competitive collaboration: Joint unsupervised learning of depth,
  camera motion, optical flow and motion segmentation.
\newblock In {\em CVPR}, pages 12240--12249, 2019.

\bibitem{rebuffi2018efficient}
S.-A. Rebuffi, H.~Bilen, and A.~Vedaldi.
\newblock Efficient parametrization of multi-domain deep neural networks.
\newblock In {\em CVPR}, pages 8119--8127, 2018.

\bibitem{wang2018learning}
C.~Wang, J.~Miguel~Buenaposada, R.~Zhu, and S.~Lucey.
\newblock Learning depth from monocular videos using direct methods.
\newblock In {\em CVPR}, pages 2022--2030, 2018.

\bibitem{wang2018non}
X.~Wang, R.~Girshick, A.~Gupta, and K.~He.
\newblock Non-local neural networks.
\newblock In {\em CVPR}, pages 7794--7803, 2018.

\bibitem{wang2004image}
Z.~Wang, A.~C. Bovik, H.~R. Sheikh, and E.~P. Simoncelli.
\newblock Image quality assessment: from error visibility to structural
  similarity.
\newblock {\em IEEE TIP}, 13(4):600--612, 2004.

\bibitem{watson2019self}
J.~Watson, M.~Firman, G.~J. Brostow, and D.~Turmukhambetov.
\newblock Self-supervised monocular depth hints.
\newblock In {\em ICCV}, pages 2162--2171, 2019.

\bibitem{yang2018lego}
Z.~Yang, P.~Wang, Y.~Wang, W.~Xu, and R.~Nevatia.
\newblock Lego: Learning edge with geometry all at once by watching videos.
\newblock In {\em CVPR}, pages 225--234, 2018.

\bibitem{yin2018geonet}
Z.~Yin and J.~Shi.
\newblock Geonet: Unsupervised learning of dense depth, optical flow and camera
  pose.
\newblock In {\em CVPR}, pages 1983--1992, 2018.

\bibitem{zhou2017unsupervised}
T.~Zhou, M.~Brown, N.~Snavely, and D.~G. Lowe.
\newblock Unsupervised learning of depth and ego-motion from video.
\newblock In {\em CVPR}, pages 1851--1858, 2017.

\bibitem{zhu2020edge}
S.~Zhu, G.~Brazil, and X.~Liu.
\newblock The edge of depth: Explicit constraints between segmentation and
  depth.
\newblock In {\em CVPR}, pages 13116--13125, 2020.

\bibitem{zou2018df}
Y.~Zou, Z.~Luo, and J.-B. Huang.
\newblock Df-net: Unsupervised joint learning of depth and flow using
  cross-task consistency.
\newblock In {\em ECCV}, pages 36--53, 2018.

\end{thebibliography}

%
%
%
%
\newpage
\appendix
\section{Network Details}
\begin{figure*}[h]
    \centering
    \includegraphics[width=\linewidth]{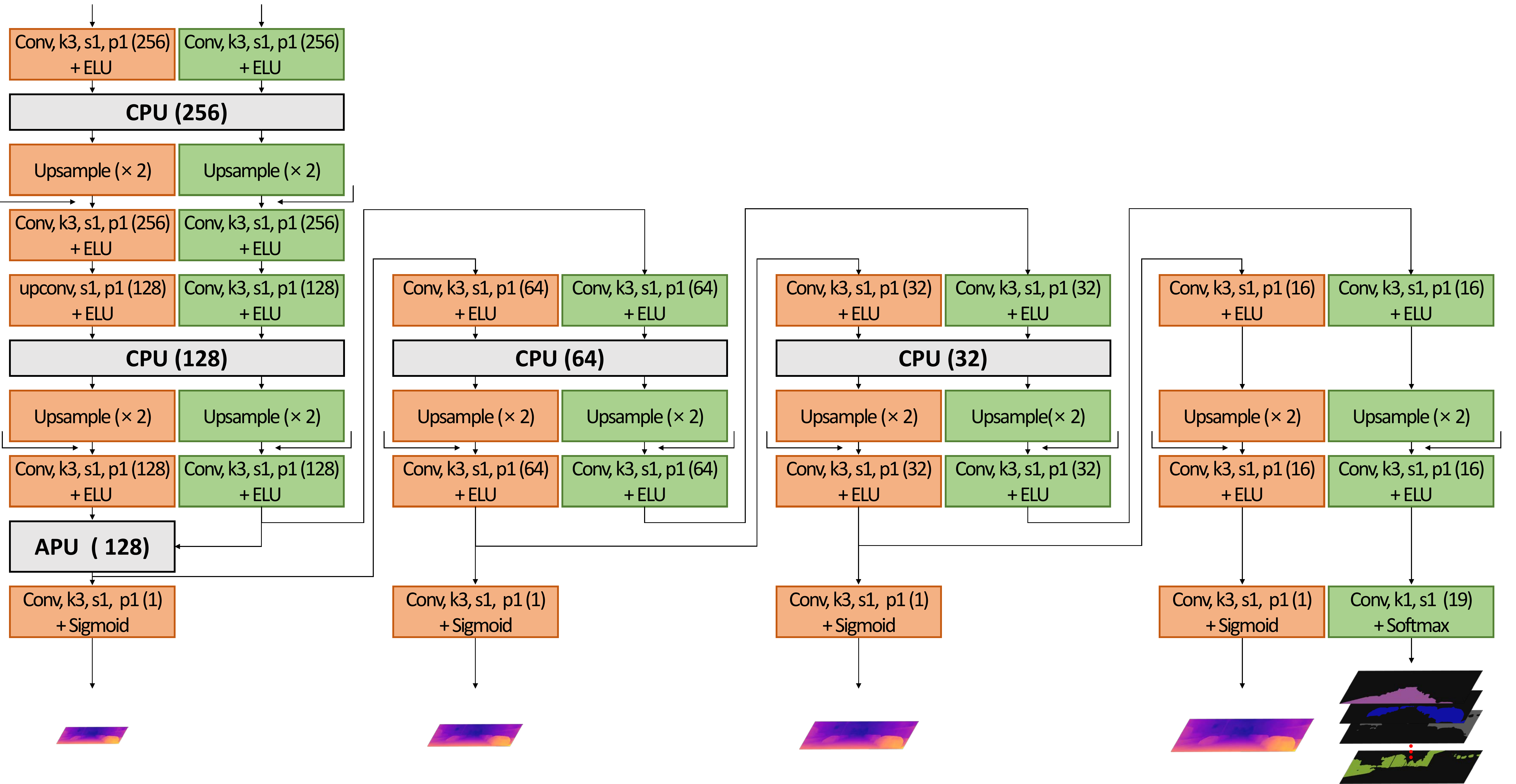}
    \caption{Details of the decoder architecture. Conv is convolutional layer, Upsample is nearest interpolation with a scale factor 2, k is kernel size, s is stride, and p is padding. 
    The numbers in the parentheses for Conv boxes denote the number of filters.
    The extrinsic arrows under the upsample boxes indicate the skip connections from the encoder features.}
    \label{network_details}
\end{figure*}

%
The details of the decoder networks are shown in Fig. \ref{network_details}.
The overall pipeline of our method consists of depth layers, semantic layers, CPU, and APU.
The orange blocks and the green blocks denote the depth layers and the semantic layers, respectively, and APU and CPU intersect the propagation of the task-specific layers to create semantic-aware depth features.
The encoder features and the corresponding decoder features with the same spatial size are concatenated along the channel dimension through the skip connections.
At the bottom of the orange blocks, there comes the multi-scale prediction of intermediate disparity maps.
The last green block gives the semantic features which have the probabilities of the classes.
%
\section{Warping Process}
Self-supervised monocular depth estimation exploits the photometric loss with SSIM \cite{wang2004image} to train the networks,
\begin{equation}
    L_{photo} = \frac{1}{N}\sum_{p\in N}{(\alpha \frac{1-\text{SSIM}_{mn}(p)}{2}
              + (1-\alpha)\parallel I_{m}(p)-I^\prime_{m}(p) \parallel)}\:.
\label{photometric_loss}
\end{equation}
To obtain $I^\prime_{m}$ in the coordinate system of $I_{m}$, warping the image $I_{n}$ into the image plane of $I_{m}$ is required.
The warping process is different depending on the type of input images, which can be either left-right stereo pairs or video sequences.
Since stereo type inputs include a rectified left-right pair, which is pre-calibrated and aligned on the same image plane, only the difference in the x-direction is considered through the disparity map.
Therefore, when $I_{m}$ and $I_{n}$ are left and right images, the equation of image reconstruction is as follows:
\begin{equation}
    I^\prime_{m}(p) = I_{n}(p-d^{l}_{m}(p))\:,
\end{equation}
where $p$ indicates the pixel coordinate and $d^{l}_{m}$ denotes the left disparity map predicted from $I_{m}$.
Conversely, to synthesize the left image from the right image, it is required to swap $m$ and $n$ in Eq. \ref{photometric_loss}, and the expression is as follows:
\begin{equation}
    I^\prime_{n}(p) = I_{m}(p+d^{r}_{m}(p))\:,
\end{equation}
where $d^{r}_{m}$ is the right disparity map.
The predicted disparity $d$ satisfies $d = bf/D$ with the focal length $f$, the baseline distance $b$ between the cameras, and the corresponding depth map $D$.

Sequence type inputs consist of the raw data without the rectification between frames.
Therefore, in order to locate the coordinates of the frames in the same image plane, projection from $I_{n}$ to $I_{m}$'s plane using camera intrinsic matrix $K$, the estimated depth $\hat{D}$, and the relative pose $\hat{T}_{m\rightarrow n}$ is necessary. 
Let $p_{m}$ and $p_{n}$ denote the coordinates of a pixel in the frame $I_{m}$ and $I_{n}$, then we can calculate the projected pixels $p_{n}$,
\begin{equation} 
    p_n \sim  K\hat{T}_{m\rightarrow n}\hat{D}_{m}(p_{m})K^{-1}p_{m}\:.
\end{equation}
We assume the intrinsic parameters $K$ are the same in all the scenes; for convenience while they can be different.
As the projected coordinates are continuous in both stereo and sequence types, interpolation through the bilinear sampling is needed following the spatial transformer networks \cite{jaderberg2015spatial}. Additionally, edge-aware smoothness loss \cite{godard2017unsupervised} and mean-normalized inverse depth \cite{wang2018learning} is used for training as follows:
\begin{equation}
    L_{smooth} = {\mid\partial_{x}d_{m}^{*}\mid}e^{-{\mid\partial_{x}I_{m}\mid}} + {\mid\partial_{y}d_{m}^{*}\mid}e^{-{\mid\partial_{y}I_{m}\mid}},
\label{smooth_loss}
\end{equation}
\begin{equation}
    d_{m}^{*} = d_{m} / {\overline{d_{m}}}.
\end{equation}

\section{Implementation Details}
We trained our model in a batch size of 8 using Adam optimizer \cite{kingma2014adam}. We used the learning rate of $10^{-4}$ and the weight decay $\beta=(0.9, 0.999)$.
The training is done end-to-end with images and precomputed segmentation masks resized to 640 $\times$ 192 (512 $\times$ 256 for stereo). We set \(\lambda_{seg} = 1\) and \(\lambda_{smooth} = 10^{-3}\) to balance the loss function. The remaining details follow \cite{godard2019digging}, which is our method's base network. All depth estimation performance was measured on an NVIDIA GTX 2080Ti GPU.

\section{Self-supervised Models based on Stereo Images}
In stereo model, we used Eigen \cite{eigen2014depth}'s splits of 22,600 left-right pairs for training and 888 pairs for validation. The test split is composed of 697 images. In order to demonstrate the scalability of our method in self-supervised monocular depth estimation, the proposed modules are applied to Monodepth2, which train the networks from stereo cues. Table \ref{stereo} shows that semantic-aware depth features in the stereo model also increase the performance comparable to recent methods \cite{chen2019towards, watson2019self}, which only focus on self-supervised training with stereo vision. 
On the other hand, our method can be globally adjusted to self-supervised networks regardless of stereo or sequence input. Moreover, \cite{watson2019self} exploit proxy disparity labels obtained by Semi-Global Matching (SGM) algorithms \cite{hirschmuller2005accurate} as additional pseudo ground truth supervision. The stereo proxy labels can be a strong supervision to boost self-supervised depth estimation performance. Hence, we expect better performance if techniques proposed by either \cite{chen2019towards} or \cite{watson2019self}  is applied to our method.

\begin{table*}[h]
    \centering
    \resizebox{\textwidth}{!}{\normalsize\begin{tabular}{l|cccc||cccc|ccc}
    \hline
    \multirow{2}{*}{Model} & \multirow{2}{*}{Seg} & \multirow{2}{*}{R/N} & \multirow{2}{*}{CPU} & \multirow{2}{*}{APU} & \multicolumn{4}{c|}{Lower is better.} & \multicolumn{3}{c}{Higher is better.} \\
    & & & & & Abs Rel & Sq Rel\: & RMSE & RMSE log & $\delta<1.25$ & $\delta<1.25^2$ & $\delta<1.25^3$\\ 
    \hline
    Garg \textit{et al.} \cite{garg2016unsupervised}* &  &  &  &  & 0.152 & 1.226 & 5.849 & 0.246 & 0.784 & 0.921 & 0.967\\
    Monodepth \cite{godard2017unsupervised}* &  &  &  &  & 0.133 & 1.142 & 5.533 & 0.230 & 0.830 & 0.936 & 0.970\\
    3Net \cite{poggi2018learning} &  &  &  &  & 0.129 & 0.996 & 5.281 & 0.223 & 0.831 & 0.939 & 0.974 \\
    Chen \textit{et al.} \cite{chen2019towards} + pp & \checkmark &  &  &  & 0.118 & 0.905 & 5.096 & 0.211 & 0.839 & 0.945 & 0.977\\
    Monodepth2 \cite{godard2019digging} &  &  &  &  & 0.109 & 0.873 & 4.960 & 0.209 & 0.864 & 0.948 & 0.975\\
    Watson \textit{et al.} \cite{watson2019self} + pp &  &  &  &  & \bf{0.106} & 0.780 & 4.695 & 0.193 & 0.875 & 0.958 & 0.980\\
    \hline
    SAFENet $(640\times192)$ & \checkmark &  &  &  & 0.125 & 0.940 & 5.049 & 0.209 & 0.853 & 0.950 & 0.977\\
    SAFENet $(640\times192)$ & \checkmark & \checkmark &  &  & 0.118 & 0.888 & 4.919 & 0.201 & 0.868 & 0.954 & 0.978\\
    SAFENet $(640\times192)$& \checkmark & \checkmark & \checkmark & \checkmark & 0.114 & 0.799 & 4.708 & 0.191 & 0.874 & 0.959 & 0.982\\
    \textbf{SAFENet} $(640\times192)$ + pp & \checkmark & \checkmark & \checkmark & \checkmark & 0.113 & \bf{0.775} & \bf{4.644} & \bf{0.189} & \bf{0.877} & \bf{0.960} & \bf{0.982}\\
    \hline
    SAFENet $(1024\times320)$ & \checkmark & \checkmark & \checkmark & \checkmark & 0.111 & 0.773 & 4.613 & 0.188 & 0.878 & 0.960 & 0.982\\
    \textbf{SAFENet} $(1024\times320)$ + pp & \checkmark & \checkmark & \checkmark & \checkmark & 0.110 & \bf{0.751} & \bf{4.553} & \bf{0.187} & \bf{0.880} & \bf{0.961} & \bf{0.982}\\
    \hline
    \end{tabular}}
    \caption{Ablation for stereo model. The term pp means the post-processing method \cite{godard2017unsupervised}.
    }
    \label{stereo}
\end{table*}

\section{Additional Experimental Results}

\begin{figure*}[h]
    \centering
    \includegraphics[width=\linewidth]{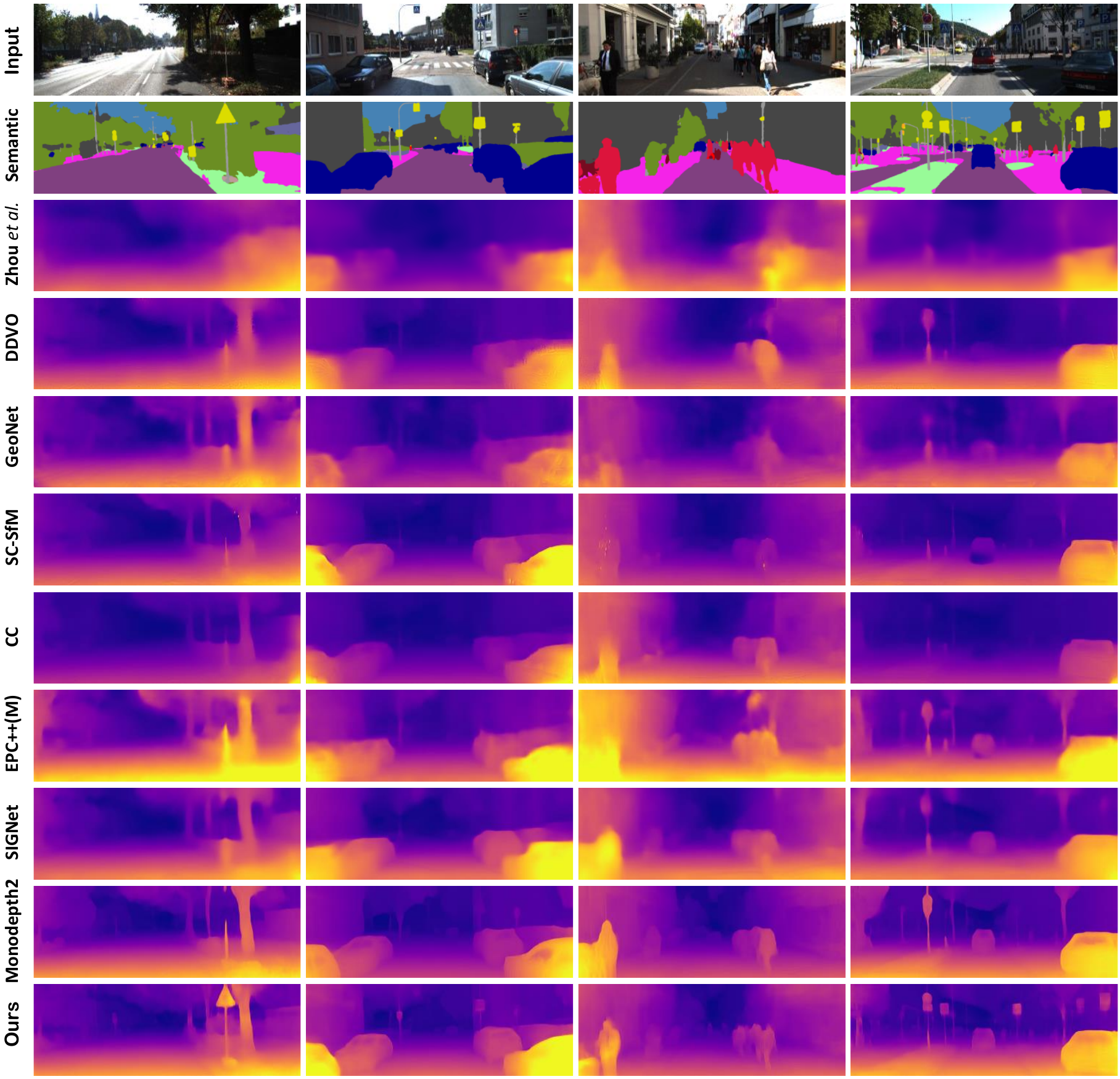}
    \caption{Qualitative results on the KITTI Eigen split. Our models in the last row produce better visual outputs, especially the sharpest boundaries of the objects.
    In the second row, Semantic denotes the segmentation results from DeepLabv3+ \cite{chen2018encoder} on the test set. 
    }
    \label{depth_visual1}
\end{figure*}

\begin{figure*}[t]
\begin{center}
    \centering
    \includegraphics[width=\linewidth]{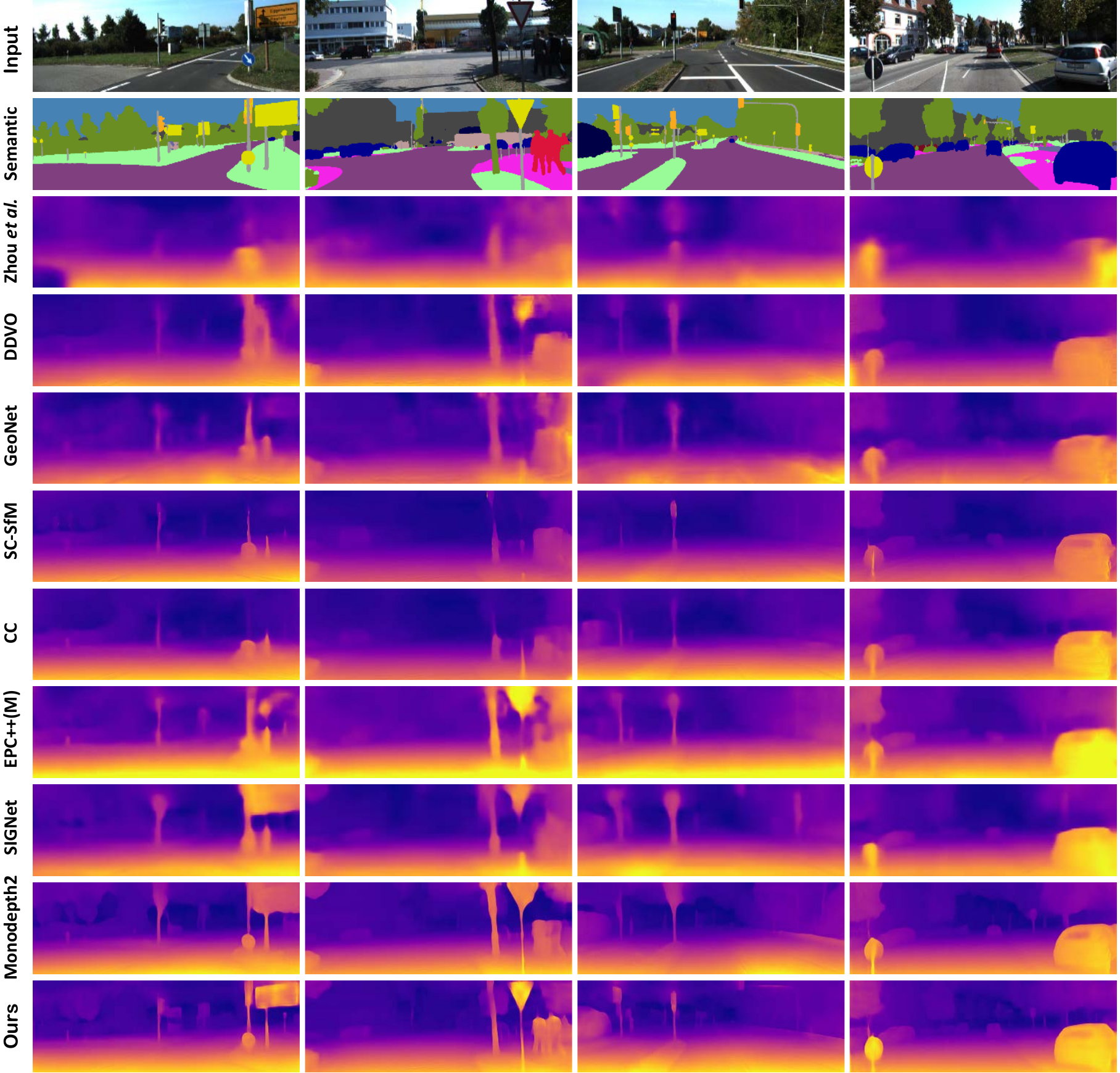}
    \caption{Qualitative results on the KITTI Eigen split.}
    \label{depth_visual2}
\end{center}
\end{figure*}

\begin{figure*}[h]
    \centering
    \includegraphics[width=\linewidth]{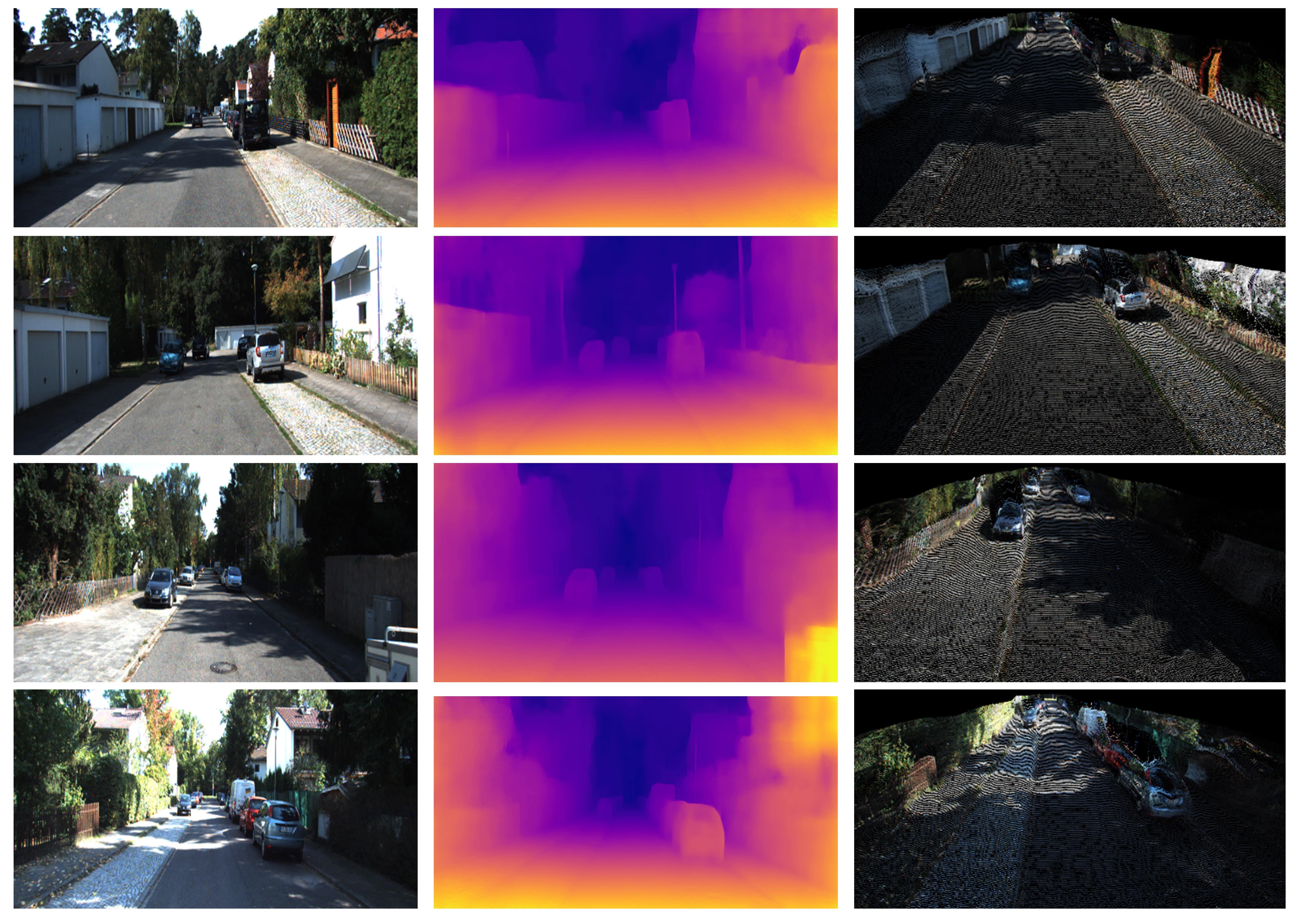}
    \caption{Examples of our reconstructed point clouds based on self-supervision from monocular video sequences.}
    \label{Reconst_pcds}
\end{figure*}

In Fig \ref{depth_visual1} and Fig \ref{depth_visual2}, we show additional qualitative comparison with other networks for the KITTI Eigen split. In addition, we show the 3D point cloud reconstructed from the predicted depth map in Fig \ref{Reconst_pcds}. More experimental results of reconstructed point clouds can be found on the supplementary video attached. 
The qualitative results in Fig. \ref{depth_visual2} show that our approach reduces the problem that training with photometric losses is inappropriate to where ambiguous boundaries or complicated shapes exist.
For example, road signs in the first and last columns are the hard objects to describe, so all the other methods except ours fail to estimate the depth accurately.
As our method with semantic-aware depth features perceives the representation of the target objects, the outlines of instances become clear.

\begin{figure*}[]
    \centering
    \includegraphics[width=\linewidth]{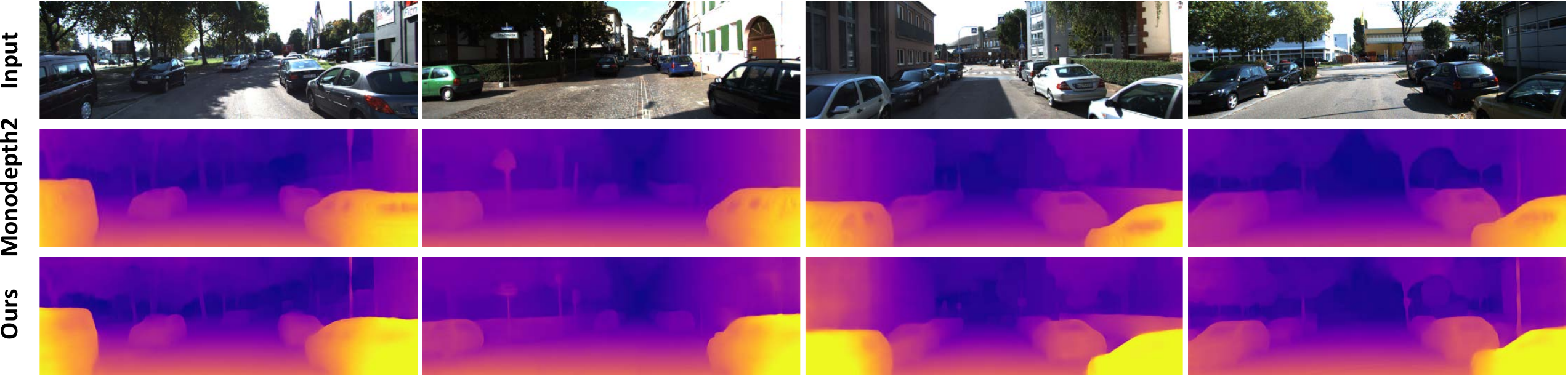}
    \caption{Reflective material examples. Ours estimates relatively consistent depth values with the surroundings, even in the areas where Lambertian assumptions are ignored. }
    \label{Lambertian}
\end{figure*}

\begin{figure*}[]
    \centering
    \includegraphics[width=0.8\linewidth]{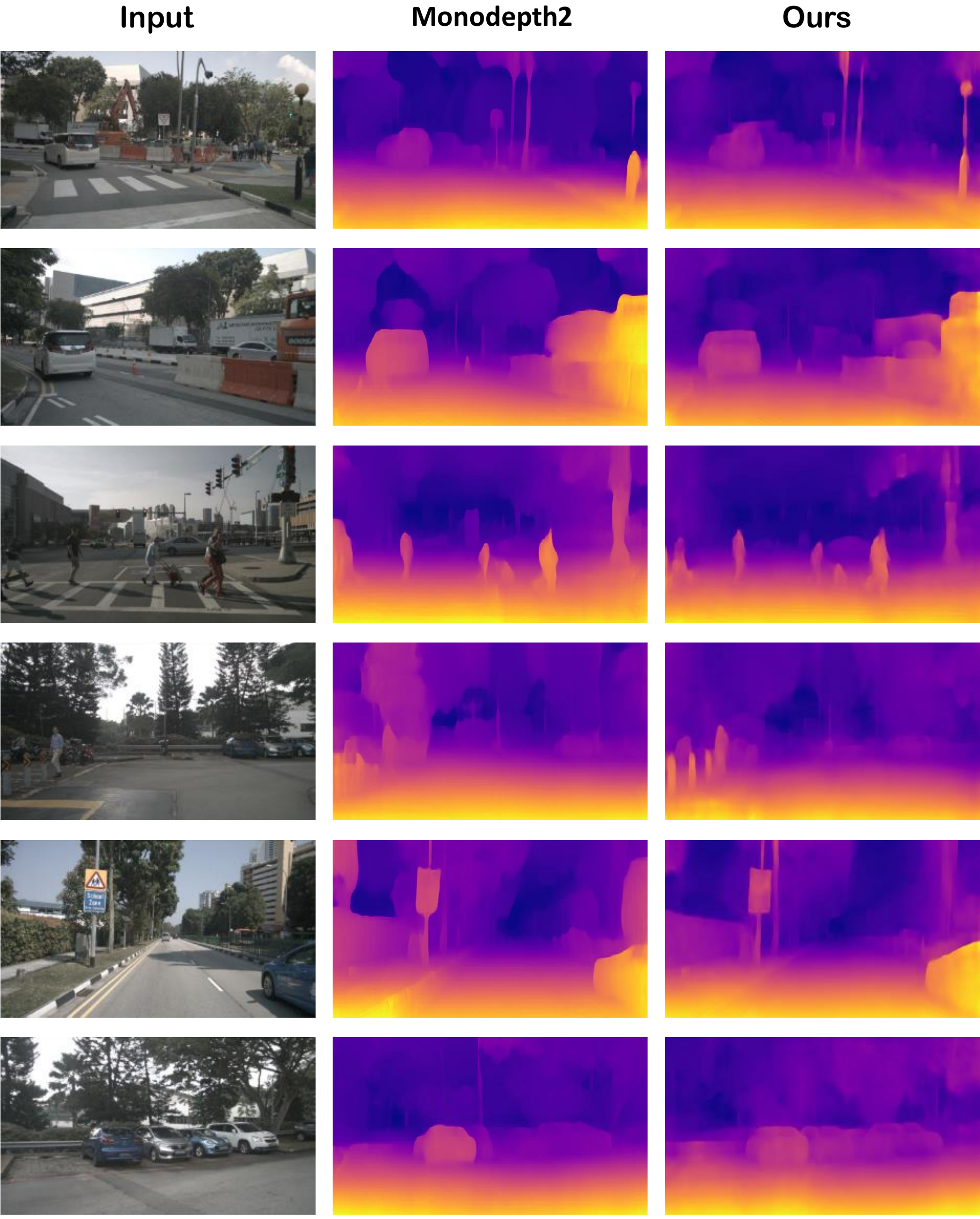}
    \caption{nuScenes qualitative results}
    \label{nuScenes}
\end{figure*}

Figure \ref{Lambertian} demonstrates that our approach has better qualitative results in the regions where the Lambertian assumption is violated.
Without semantic-awareness, Monodepth2 \cite{godard2019digging} often fails to learn proper depths for distorted, reflective, or color-saturated regions like windows of vehicles. However, our model is aware of semantic information which can tell whether a group of neighboring pixels belongs to the same object category or not. Therefore, the distances of the windows are similar to those of their vehicles compared to \cite{godard2019digging}. In Fig \ref{nuScenes}, We demonstrate the qualitative results on nuScenes dataset to show the generalization capability.

\end{document}